\documentclass{article}

\PassOptionsToPackage{compress, sort}{natbib}

\usepackage[final]{neurips_2018}

\usepackage[utf8]{inputenc} 
\usepackage[T1]{fontenc}    

\usepackage{amsfonts}
\usepackage{amssymb}
\usepackage{amsmath}
\usepackage{amsthm}
\usepackage{dsfont}
\usepackage{pbox}
\usepackage{calrsfs}

\usepackage{bm}
\usepackage{cancel}
\usepackage{bbold}
\usepackage{bm}
\usepackage{slashed}
\usepackage{color}
\usepackage{algorithm}
\usepackage{algpseudocode}
\usepackage{tikz}

\usepackage{hyperref}       
\usepackage{url}            
\usepackage{booktabs}       
\usepackage{nicefrac}       
\usepackage{microtype}      
\usepackage{subcaption}
\usepackage{bm}

\newcommand{\mc}{\mathcal}

\DeclareMathOperator*{\argmin}{arg\,min}

\title{Recurrent machines for likelihood-free inference}

\author{
  Arthur Pesah\thanks{Both authors contributed equally to this work.} \\
  KTH Royal Institute of Technology\\
  Stockholm, Sweden \\
   \And
   Antoine Wehenkel\footnotemark[1] \\
   University of Liège \\
   Liège, Belgium \\
   \And
   Gilles Louppe \\
   University of Liège \\
   Liège, Belgium \\
}

\DeclareMathOperator{\diag}{diag}

\begin{document}

\maketitle

\begin{abstract}
Likelihood-free inference is concerned with the estimation of the parameters of a non-differentiable stochastic simulator that best reproduce real observations. 
In the absence of a likelihood function, most of the existing inference methods optimize the simulator parameters through a handcrafted iterative procedure that tries to make the simulated data more similar to the observations. 
In this work, we explore whether meta-learning can be used in the likelihood-free context, for learning automatically from data an iterative optimization procedure that would solve likelihood-free inference problems.
%
We design a recurrent inference machine that learns a sequence of parameter updates leading to good parameter estimates, without ever specifying some explicit notion of divergence between the simulated data and the real data distributions.
We demonstrate our approach on toy simulators, showing promising results both in terms of performance and robustness.

\end{abstract}

\section{Introduction}
Modern science often relies on the modeling of complex data generation process by means of computer simulators. While the forward generation of observables is often straightforward and well-motivated, inverting  generation processes is usually very difficult. In particular, scientific simulators are often stochastic and give rise to intractable likelihood functions that prevent the use of classical inference algorithms. 
The importance and prevalence of this problem has recently motivated the development of so-called likelihood-free inference methods (LFI) which do not make use of the likelihood function for estimating model parameters. 
LFI methods~\citep[e.g.,][]{avo, abc, bolfi} are often based on handcrafted iterative optimization procedures, where a sequence of updates are performed to make the simulated data more similar to the observations.


Driven by the promises of learning to learn, meta-learning has shown that automatically learning neural optimizers from data is possible, achieving results close to the state-of-the-art for the task of training neural networks~\citep{learning2learn} or solving inverse problems~\citep{rim} when the gradient of the objective function is available. Meanwhile, \citep{learning2learnwg} have shown that meta-learning is also capable of learning neural optimizers that rely at each step on the value of the objective function only, without requiring access to its gradient. 



In this work, we push the limits of the meta-learning framework further by showing that it can be used even when no explicit objective value is available at each step. More specifically, we focus on likelihood-free inference problems and build a recurrent inference machine that learns an iterative procedure for updating simulator parameters such that they converge (in distance) towards nominal parameter values known at training. In particular, the inference machine is never given access to an explicit objective function that would estimate some divergence between the synthetic distribution and the real data distribution. Rather, both the optimization procedure and the implicit objective to minimize are learned end-to-end from artificial problems.






\section{Problem statement}



The goal of likelihood-free inference is the estimation of the parameters of a stochastic generative process with an unknown or intractable likelihood function. 
Formally, given a set of observations $X^r=\{\bm{x}^r_1,...,\bm{x}^r_{M^r}\}$  drawn i.i.d. from the real data distribution $p_r(\bm{x})$, we are interested in finding the model parameters $$\bm{\theta}^* = \argmin_\theta \rho(p_r(\bm{x}), p(\bm{x}| \bm{\theta}))$$ that minimize some discrepancy $\rho$ between the real data distribution $p_r(\bm{x})$ and the implicit distribution $p(\bm{x}|\bm{\theta})$ of the simulated data. 

In this work, instead of defining one objective point $\bm{\theta}^*$, we target a probability distribution over the $\bm{\theta}$ space. Formally, our goal is to find the optimal parameter $\bm{\psi}^*$ of a parametric proposal distribution $q(\bm{\theta}|\bm{\psi})$ on $\bm{\theta}$, leading to the following optimization problem 
$$\bm{\psi}^* = \argmin_{\bm{\psi}} \rho(p_r(\bm{x}), p(\bm{x}|\bm{\psi})) \quad \text{where} \quad p(\bm{x}|\bm{\psi}) = \int q(\bm{\theta}|\bm{\psi}) p(\bm{x}|\bm{\theta}) \bm{d\theta}.$$ 
The proposal distribution provides two advantages: i) it leads to a variational formulation of the optimization problem that does not rely on the usage of the gradient of the function to be optimized \citep{variaoptim}, ii) by sampling $B$ points from this distribution, $\bm{\theta}_1,...,\bm{\theta}_B \sim q(\bm{\theta}|\bm{\psi})$, we get distinct parameters that we can give to the simulator. Then, by comparing the generated observations $X_i=(\bm{x_{i,1}},...,\bm{x_{i,M}})$ to the real ones $X^{r}$ for each $\bm{\theta}_i$, it is possible to figure out what the optimal update of $\bm{\psi}$ is.

Because the likelihood function $p(\bm{x}|\bm{\theta})$ cannot be evaluated, strategies must be found to use only simulated data in order to estimate $\bm{\psi}^*$. Current methods for likelihood-free inference typically rely on a handcrafted iterative update procedure where at each time step $t \in [1, T]$ the next estimate $\bm{\psi}_{t+1}$ is determined by comparing the simulated data produced from $p(\bm{x}|\bm{\psi})$ at the current parameter estimate $\bm{\psi}_t$ with the real observations $X^r$. 
In this work, our goal is to investigate whether meta-learning can be used for learning a parametrized update function $f_{\bm{\phi}}$ such that 
\begin{equation*}
    \bm{\bm{\psi}}_{t+1}=\bm{\psi}_{t} + f_{\bm{\phi}}((\bm{\psi}_1,\ldots,\bm{\psi}_t), X^r),\quad \text{with} \quad \bm{\psi}_t \to \bm{\psi}^* \quad \text{as} \quad t \to \infty.
\end{equation*}




\section{Recurrent machines for likelihood-free inference}

\begin{figure}
    \centering
    \includegraphics[width=1\textwidth]{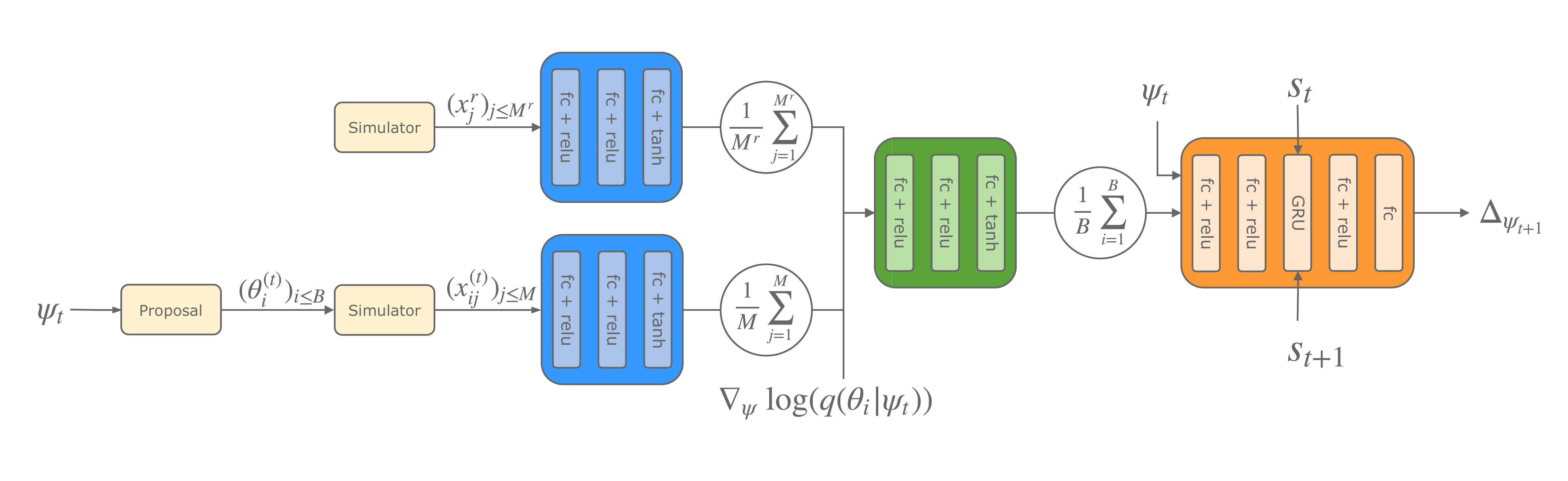}
    \caption{Recurrent machine for likelihood-free inference. All three encoders are fully-connected (fc) neural networks. The RNN is a GRU \citep{gru} with fully-connected pre-processing and post-processing layers.}
    \label{fig:archi-alfi}
\end{figure}

In this work, we define $f_{\bm{\phi}}$ as a recurrent neural network (RNN) whose architecture is given in Figure \ref{fig:archi-alfi}. At each time step $t$, the RNN takes as input the current proposal parameters $\bm{\psi}_t$, a memory state $s_t$, some information about the real observations and those generated with $\bm{\psi}_t$, and produces as output the parameter update $\Delta_{\bm{\psi}_{t+1}}$ and $s_{t+1}$. Observations $(x_j)_{j \leq M}$ are ingested through  data encoder architectured as a feedforward neural network that takes as input each $x_j$ independently, transforms each of them into a $d$-dimensional vector, and aggregates them through an averaging of those vectors over $j$.
It should therefore be able to compute any moment (and more general feature averages) of the distribution. If the moments are learned, they can be used to infer the parameters of the distribution by the method of matching moments, whose principle is to check compatibility between the generated data and the observed data by looking at these moments \citep{momentsLearning}. This also means that moments can capture the relevant information of a set of samples about a parameter of interest.

In our architecture, we use three encoders. The first one is for the real observations $(\bm{x}^r_j)_{j \leq M^r}$. The second one is for the generated observations at time $t$. We consider a sampling of $B$ parameters $(\bm{\theta}^{(t)}_i)_{i \leq B}$, and for each $\bm{\theta}^{(t)}_i$, the corresponding observations $(\bm{x}_{i,j}^{(t)})_{j \leq M}$. These two encoders share their weights, because the way true and generated data are summarized should be the same in order to be able to compare them. The last encoder takes the results of the first two as well as the log-likelihood value $\nabla_\psi \log q(\bm{\theta}^{(t)}_i|\bm{\psi}_t)$, as motivated by \cite{nes} and which represents the direction to follow to move the proposal distributions toward $\bm{\theta}^{(t)}_i$.


The loss used at training should encourage the network to generate updates $\Delta_{\bm{\psi}_t}$ of the proposal which yield to a final proposal $q(\bm{\theta}|\bm{\psi}_T$) that is
close to a delta-function at  $\bm{\theta}^*$.
The total loss of the RNN is expressed in a way similar to \citep{learning2learn}, i.e. as a weighted sum of a local loss evaluated for each $\bm{\psi}_t$,
\begin{equation*}
    \mathcal{L}(\bm{\psi}_T, \bm{\theta}^*) = \sum_{t=1}^T w_t \ell(\bm{\psi}_t, \bm{\theta}^*),
\end{equation*}
where $\ell$ is a loss comparing the proposal parameters $\bm{\psi}_t$ with the real parameter $\bm{\theta}^*$, $w_t$ is a weight given to the loss at time $t$ (for instance $w_t=1$ for all $t$, or $w_t=\mathds{1}_{t=T})$. The choice of the weighting and the loss functions are discussed in the experiments section.

We call the architecture ALFI (Automatic Likelihood-Free Inference).


\section{Experiments}



To illustrate our method and compare it with a simple baseline, we performed experiments on three toy simulators, whose likelihood is known and consequently for which the maximum likelihood estimator (MLE) can be computed. The results for the simplest simulator is shown below and results for the two other toy problems are provided in Appendix \ref{app:supp-results}. We also tested our architecture on a simulator from particle physics, whose exact likelihood function is intractable. For this last experiment we assess the model performance by comparing the data generated at the end of the iterative process with the real observations.
The full description of each simulator is given in Appendix \ref{app:simulators}.

\subsection{Illustrative example}

As an illustrative example, we implemented a simulator that generates samples from a Poisson distribution $\mathcal{P}(\lambda=e^{\theta})$. The goal is to estimate the parameter $\theta$ corresponding to real samples of this distribution.


\paragraph{Results}

\begin{figure}[htbp]
    \centering
    \begin{minipage}[t]{.47\textwidth}
        \includegraphics[width=1.\textwidth]{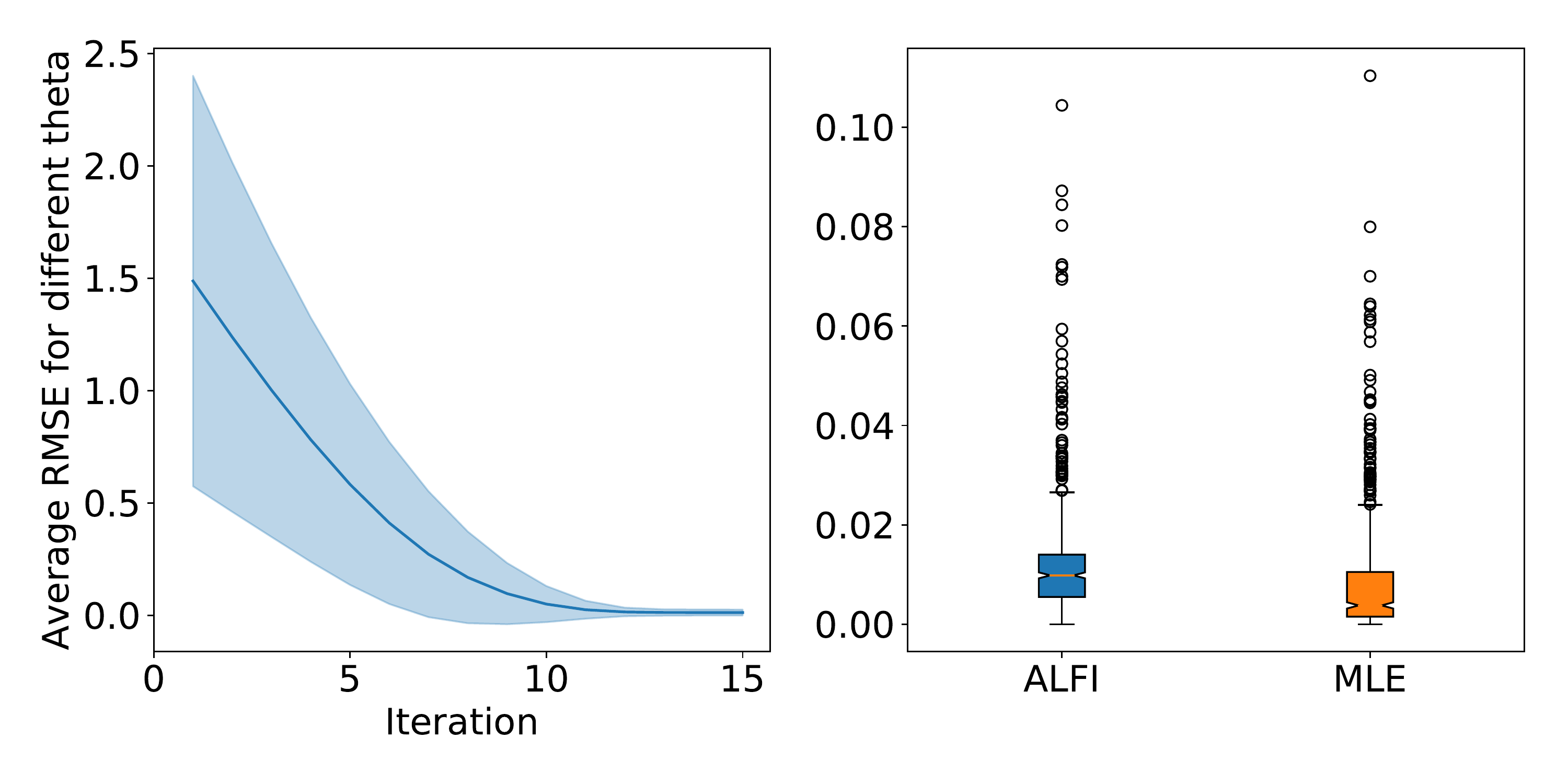}
        
    \end{minipage}
    \vspace{-1em}
    \caption{Results for the Poisson simulator. \textbf{(Left)} We observe that the RMSE decreases quickly during the 15 first iterations. \textbf{(Right)} ALFI and MLE results are very similar.}
    \label{fig:poisson}
\end{figure}

The box plot of Figure \ref{fig:poisson} compares the performance of our model with the MLE. For our model (ALFI), the root mean-squared error (RMSE) is computed between the mean of the final proposal parametrized by $\psi_{T}$ and the true value of the parameters: $\textrm{RMSE}=||\theta^* - \mathbb{E}_{\theta \sim q(\theta|\psi_T)}[\theta]||_2$. 
We observe that our model achieves similar performance as the MLE whereas it is not given any explicit function to minimize during testing.

The left part of Figure \ref{fig:poisson} shows the evolution of the average and the standard deviation of the RMSE along the iterative procedure over all test problems. It can be observed from this plot that our model quickly converges to a proposal distribution with an expected value close to $\theta^*$.

\begin{figure}
    \centering
    \includegraphics[width=.21\textwidth]{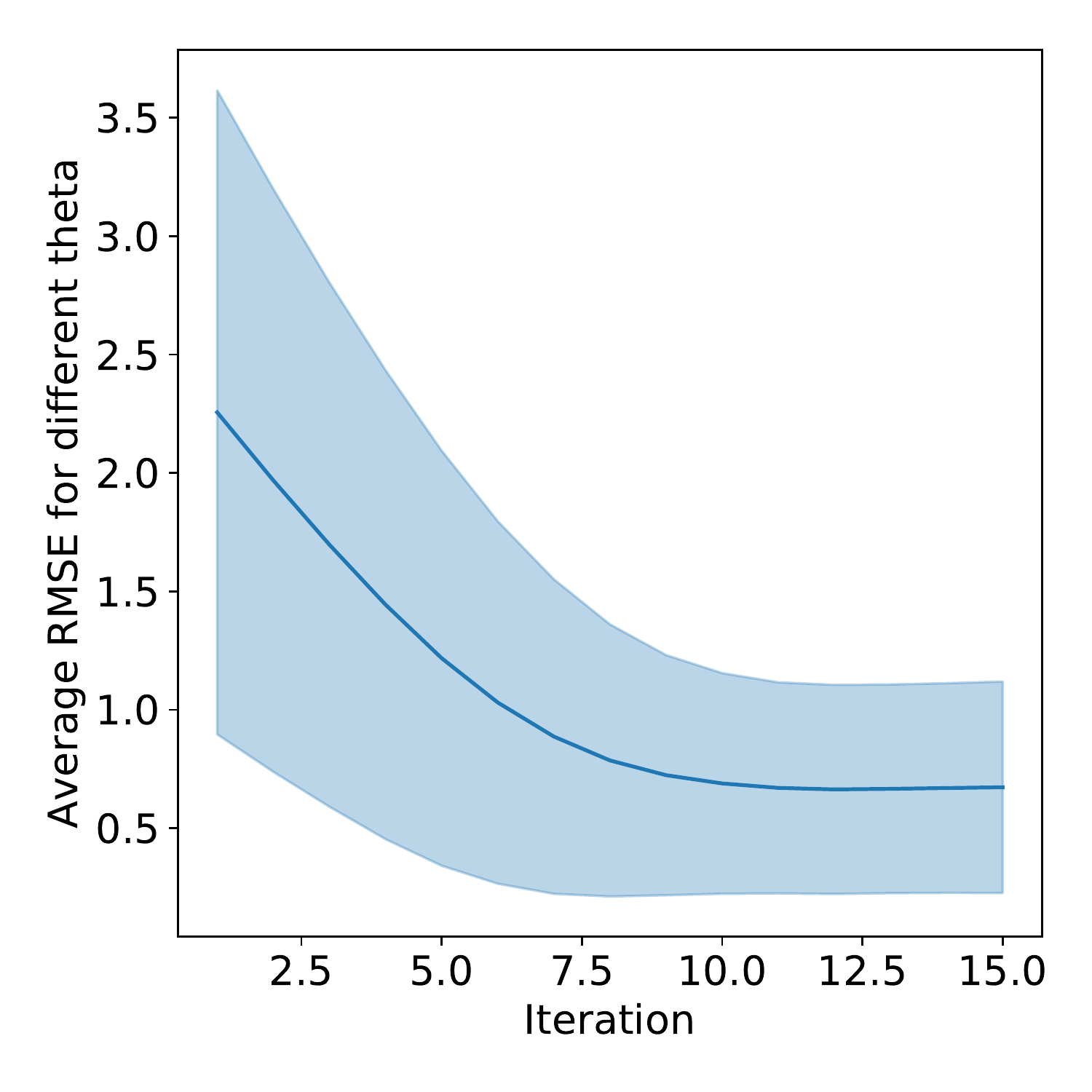}
    \includegraphics[width=.65\textwidth]{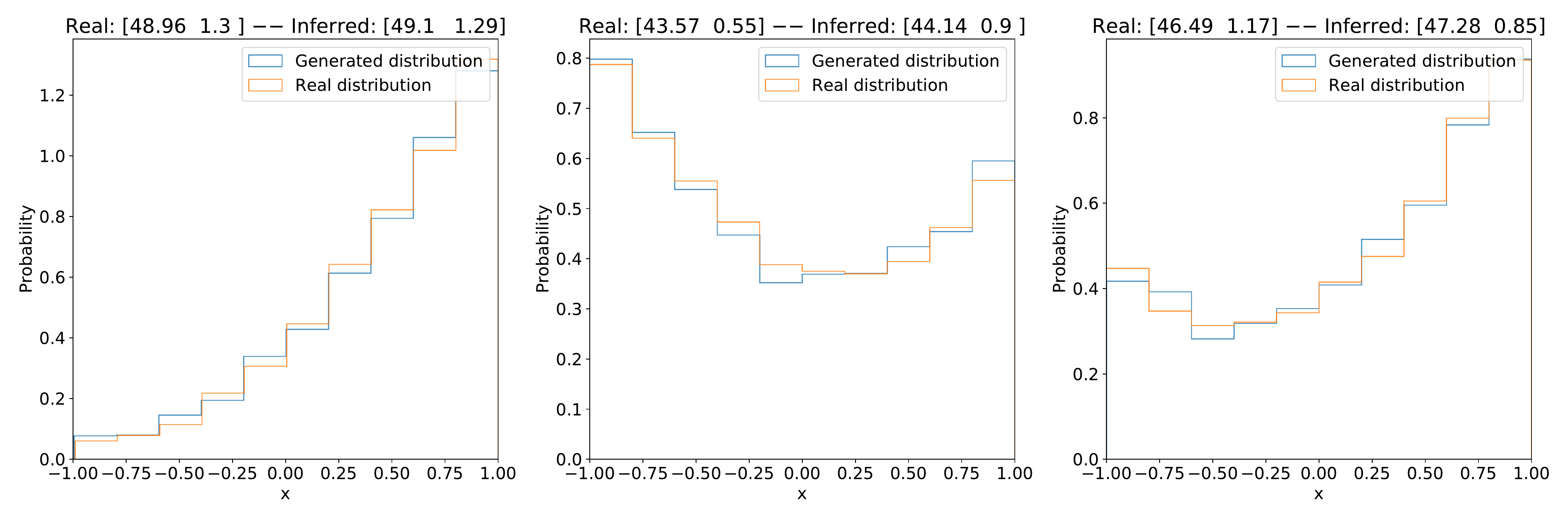}
    \caption{\textbf{(Left)} Evolution of the error between the true and predicted parameters for the Weinberg simulator, averaged over several $\theta^*$.
    \textbf{(Right)} Histogram of the real VS. generated data with the inferred parameters, for 3 different $\bm{\theta}^*$. The vectors on top of each figure represent the real parameters $[\theta^*_0,\theta^*_1]$ (left) and the inferred ones $[\hat{\theta}_0,\hat{\theta}_1]$}
    \label{fig:weinberg}
\end{figure}

\subsection{(Simplified) particle physics simulation}

We also tested our model on a simplified simulator from particle physics, called Weinberg and introduced as a likelihood-free inference benchmark in \citep{avo}. This benchmark comes with two parameters (the beam energy and the Fermi constant) and one observable (the cosine of the scattering angle). 

\paragraph{Results} Figure \ref{fig:weinberg} (left) presents the evolution of the error between the real and the predicted parameters along 15 iterations. We see that it has learned an iterative procedure and converges after 10 iterations. Figure \ref{fig:weinberg} (right) compares the distributions of the simulated data $p(x|\bm{\psi}_T)$ and the real data $p_r(x)$. We can see that ALFI has managed to infer parameters that simulate a realistic distribution.

\subsection{Robustness of the model}

In order to evaluate the robustness of the learned optimization procedure, we tested the model on a number of iterations $T_{\textrm{test}}$ greater than the number $T_{\textrm{train}}$ used during training. On the Poisson and the multivariate simulators (Appendix \ref{app:cmd}), we observed that increasing the number of iterations improves the performance of ALFI. Therefore, our model seems to 
learn an update rule $\bm{\Delta}_\psi$ that is not tied to the specific number of updates used a training, but rather generalizes to a larger horizon.
Those results are shown in Appendix \ref{app:supp-results}.





\section{Conclusions and future works}
In this work, we provide a proof-of-concept of a meta-learning architecture designed to solve likelihood-free inference problems. We applied our model on toy simulators and achieved results competitive with maximum likelihood estimation. We finally applied it on a simple particle physics simulator, and showed that it can infer parameters corresponding to samples close to the real ones.

As for future work, we see two paths worth of exploration. First, getting a better understanding of the optimization procedure learned by ALFI: can we interpret the representation learned by each data encoder? Could an update model learned for a simulator be transferred to another simulator? Is the learned procedure comparable to other existing methods? Secondly, evaluating our model on more complex simulators and comparing it to state-of-the-art LFI methods: since our approach is intensive in the number of simulator calls, moderating this complexity would be a necessary step to scale our method to slower simulators.


\newpage
\bibliographystyle{apalike}
\bibliography{bibliography}

\clearpage
\appendix

\section*{Appendix}


\section{Experimental setup}
\paragraph{Optimizer} To train our architecture, we used the ADAM optimizer \citep{adam}.

\paragraph{Proposal distribution} For all the experiments, we defined $q(\bm{\theta}|\bm{\psi})$ as a multivariate Gaussian distribution $\mathcal{N}(\bm{\mu}, \diag(\bm{\sigma}))$ with parameters $\bm{\psi} \in \mathbb{R}^{d}$ where $d$ denotes the size of the parameter space and $\bm{\psi}_i=[\bm{\mu}, \bm{\sigma}] \quad \text{where} \quad \bm{\mu}, \bm{\sigma} \in \mathbb{R}^{d}$. For each experiment the initial proposal $\bm{\psi}_1$ is made of a random mean vector $\bm{\mu}_1$ drawn with the same distribution as the parameters $\bm{\theta}^*$ of the different problems of the training set. The variance vector $\bm{\sigma}_1$ always starts at $\bm{\sigma}_1 = [e^{0.5}, ..., e^{0.5}]$.

\paragraph{Partial loss function} For the partial loss $\ell(\bm{\psi}_t,\bm{\theta}^*)$ at time t, we tried two distinct functions: the mean-squared error between the mean of $\bm{\psi}_t$ and $\bm{\theta}^*$, $\ell(\bm{\psi}_t,\bm{\theta}^*)=|| \mathbb{E}_{\bm{z}\sim q(\bm{\theta}|\bm{\psi})}[\bm{z}] - \bm{\theta}^* ||^2_2$ and the negative-log-likelihood of the proposal evaluated on $\bm{\theta}^*$, $\ell(\bm{\psi}_t,\bm{\theta}^*)=-\log(q(\bm{\theta}^*|\bm{\psi}))$. The later loss has the advantage of taking more the variance of the proposal into account and lead to better performance experimentally. Therefore, we decided to use the likelihood loss function for all the subsequent experiments.

\paragraph{Weighting function} The choice of the weighting function $w_t$ determines the exploration-exploitation trade-off of our iterative algorithm. We tested three weighting schemes: 
\begin{itemize}
    \item $w_t=\mathds{1}_{t=T}$: only $\bm{\psi}_T$, the final proposal distribution on the parameters, is taken into account in the total loss function. It means that the algorithm can freely explore the parameter-space during $T-1$ iterations.
    \item $w_t=1$ for all $t$: all the parameters' estimates found during the iterative process are taken into account with the same weight. It encourages the algorithm to converge as fast as possible to a good parameter.
    \item $w_t=\frac{e^{\beta x}-1}{e^{\beta}-1}$: compromise between the two previous weightings. The first steps are given a low weight, encouraging exploration, while the last steps have a high weight to ensure convergence by the end.
\end{itemize}

Among those three weighting schemes, the exponential one gave the best performance and we decided to use it for all the subsequent experiments.

\paragraph{Marginalization}
To avoid overfitting, the initial value of the mean of the proposal parameters is taken randomly. Thus to compute the performance of our model at test time, $\bm{\psi}_1$ is marginalized out. To do so, we draw 500 $\bm{\psi}_1$ values and take the average outputs $\bm{\psi}_T$.

\paragraph{Hyperparameters} We provide below a list of all the hyperparameters of ALFI, along with a description if necessary:

\begin{itemize}
    \item Number of epochs
    \item Number of iterations $T$
    \item Number of $\bm{\theta}^*$ for meta-training: size of the meta-dataset
    \item Distribution of the $\bm{\theta}^*$: how the $\bm{\theta}^*$ used in the meta-training are generated
    \item Meta batch-size: number of $\bm{\theta}^*$ that we use to compute the gradient that we backpropagate in our networks.
    \item Batch size for $\bm{\theta}$: number of $\bm{\theta}$ that we generate from $q(\bm{\theta}|\bm{\psi}_t)$ at each time $t$
    \item Batch size for $\bm{x}$: number of $\bm{x}$ that we generate from each $\bm{\theta}$ generated from $\bm{\psi}_t$ at time $t$
    \item Learning rate
    \item Clipping: we force our model to follow an iterative procedure by clipping each component of the output $\bm{\Delta}_\psi$ of the RNN between two values.
\end{itemize}


\section{Simulators} \label{app:simulators}

\subsection{Linear Regression}
\paragraph{Forward generation} The data generated by this procedure follow a linear law in 2 dimensions, the unknown parameters of the simulator represents the slope and the offset of this line. Formally, let $X = [x, 1, y]$ denote an observation generated by $\theta$, where $x, y \in \mathbb{R}$. Then X satisfies the constraint $y - n = \tan(\theta_0)x + \theta_1$ with $n\sim \mc{N}(0, 0.1)$ and the value of $x$ being drawn uniformly between $-1$ and $1$.

\paragraph{Hyperparameters}
\begin{itemize}
    \item Number of epochs: $300$
    \item Number of iterations $T$: $15$ 
    \item Number of $\bm{\theta}^*$ for meta-training: $10000$
    \item Distribution of the $\bm{\theta}^*$: uniformly in $[0,\frac{\pi}{2}]\times[-1, 1]$
    \item Meta batch-size: $16$
    \item Batch size for $\bm{\theta}$: $20$
    \item Batch size for $\bm{x}$: $20$
    \item Learning rate: $1e-3$
    \item Clipping: $[-0.25,0.25]$
\end{itemize}

\subsection{Poisson distribution}
\paragraph{Forward generation} The forward generation process is a simple Poisson distribution which depends on the mean parameter $\lambda \in \mathbb{R}^+$ of the distribution. To make the parametrisation real we define $\bm{\theta} = \theta = \log(\lambda) \in \mathbb{R}$. The generation of an observation $x$ conditionally to the parameter value $\theta$ is done by sampling $x$ from $\mc{P}(\lambda=e^\theta)$.

\paragraph{Hyperparameters}
\begin{itemize}
    \item Number of epochs: $300$
    \item Number of iterations $T$: $15$ 
    \item Number of $\theta^*$ for meta-training: $10000$
    \item Distribution of the $\theta^*$: uniformly in $\left[0.2,7.0\right]$
    \item Meta batch-size: $16$
    \item Batch size for $\theta$: $20$
    \item Batch size for $x$: $20$
    \item Learning rate: $1e-3$
    \item Clipping: $[-0.5,0.5]$
\end{itemize}

\subsection{Multivariate Distribution}\label{app:cmd}
\paragraph{Forward generation} The forward generation process for $\bm{\theta} = [\theta^{(0)}, \theta^{(1)}, \theta^{(2)}]$ can be described as follow:
\begin{enumerate}
    \item Draw independently $z^{(0)} \sim \mc{N}(\theta^{(0)}, 1)$, $z^{(1)} \sim \mc{N}(3, e^{\frac{\theta^{(1)}}{3}})$, $z^{(2)} \sim \text{GMM}\left(\frac{1}{2} \mc{N}(-2, 0.5), \frac{1}{2}\mc{N}(2, 1)\right)$, $z^{(3)} \sim \mc{U}(-5, \theta^{(2)})$, $z^{(4)} \sim \text{Exp}(0.5)$
    \item Compute $\bm{x} = R\bm{z}$ with $\bm{z} = \left[z^{(0)}, z^{(1)}, z^{(2)}, z^{(3)}, z^{(4)}\right]^T$ where $R \in \mathbb{R}^{5\times 5}$ is a positive semi-definite matrix.
\end{enumerate}

\paragraph{Hyperparameters}
\begin{itemize}
    \item Number of epochs: $300$
    \item Number of iterations $T$: $15$ 
    \item Number of $\bm{\theta}^*$ for meta-training: $10000$
    \item Distribution of the $\bm{\theta}^*$: uniformly in $\left[-3,3\right]^3$
    \item Meta batch-size: $16$
    \item Batch size for $\bm{\theta}$: $20$
    \item Batch size for $\bm{x}$: $20$
    \item Learning rate: $1e-3$
    \item Clipping: $[-0.2,0.2]$
\end{itemize}

\subsection{Weinberg Simulator}

Introduced in \citep{avo}, Weinberg is a simplified simulator from particle physics of electron-positron collisions resulting in muon-antimuon pairs ($e^+ e^- \rightarrow \mu^+ \mu^-$).

\paragraph{Forward generation} The simulator takes two parameters, the Fermi constant $G^f$ and the beam energy $E^{\textrm{beam}}$, and produces the one-dimensional observable $x=\cos(A)$, where $A$ is the angle of the outgoing muon with respect to the originally incoming electron. 

\paragraph{Hyperparameters}
To generate our training dataset we draw $10^3$ parameters $[\theta_0, \theta_1]$ uniformly in the square $[40, 50]\times[0.5, 1.5]$. We enforce our model to follow an iterative procedure by clipping the step $\Delta_\psi$ output between $-0.2$ and $0.2$.

\begin{itemize}
    \item Number of epochs: $130$
    \item Number of iterations $T$: $15$ 
    \item Number of $\bm{\theta}^*$ for meta-training: $1000$
    \item Distribution of the $\bm{\theta}^*$: uniformly in the $[40,50]\times[0.5, 1.5]$
    \item Meta batch-size: $16$
    \item Batch size for $\bm{\theta} $: $8$
    \item Batch size for $x$: $64$
    \item Learning rate: $2e-4$
    \item Clipping: $[-0.2,0.2]$
\end{itemize}


\section{Supplementary results}\label{app:supp-results}

\subsection{Poisson simulator}
\begin{figure}[htbp]
    \centering
    \begin{minipage}[t]{.47\textwidth}
        \includegraphics[width=1.\textwidth]{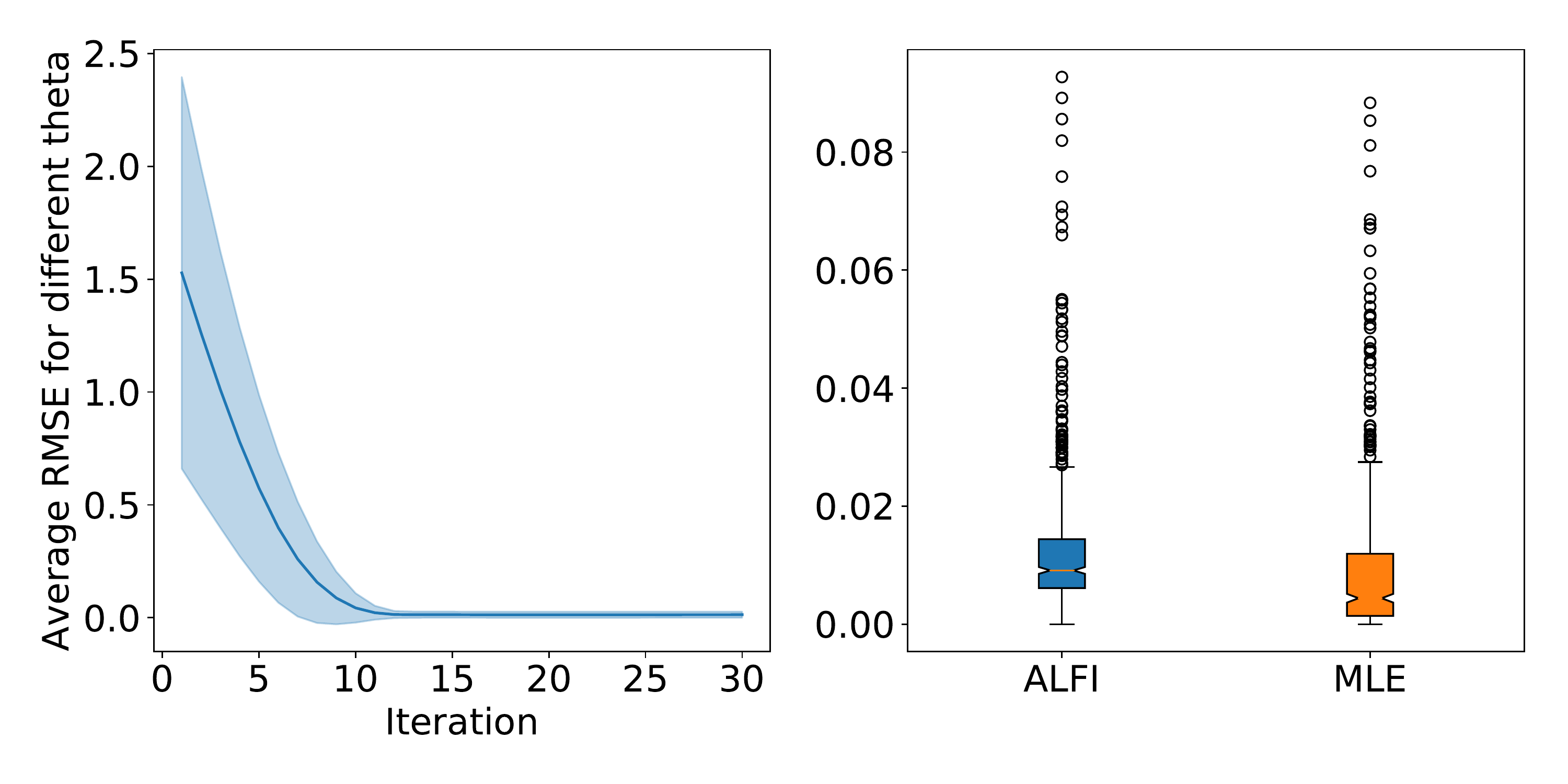}
        \caption{This plot shows that ALFI tested with a number of steps $T_{\textrm{test}}=30$ gives even better result than for $T_{\textrm{train}}=15$, in particular the upper whisker is smaller for ALFI than for MLE.}
        \label{fig:poisson-generalization}
    \end{minipage}
    \qquad
    \begin{minipage}[t]{.47\textwidth}
    \includegraphics[width=1.\textwidth]{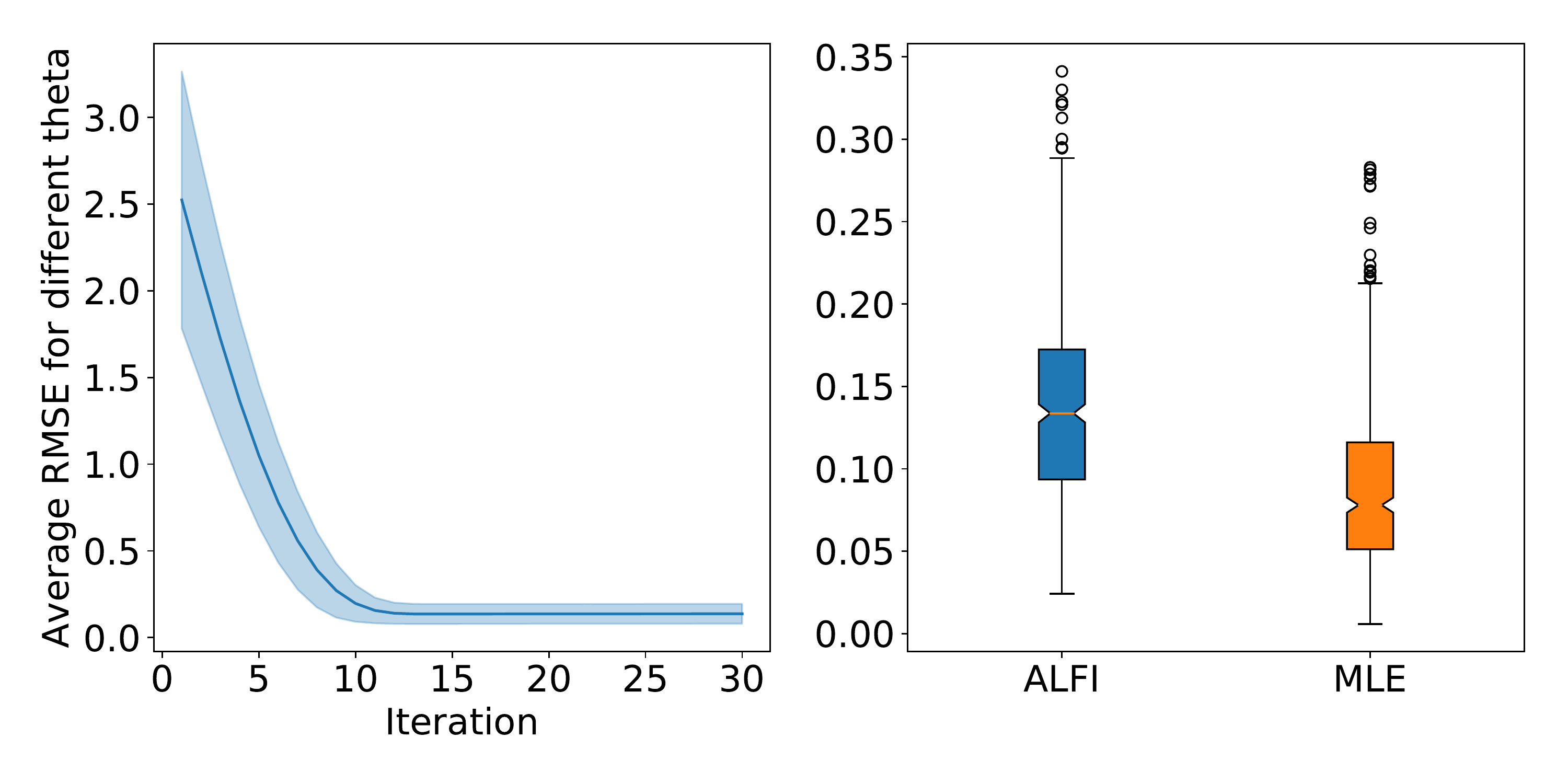}
        \caption{Results for the multivariate simulator. \textbf{(Left)} The RMSE quickly decreases during the 15 first iterations and continues to slightly decrease afterward. \textbf{(Right)} The performance of our model.}
        \label{fig:nmd}
    \end{minipage}
\end{figure}

To check that the procedure learned by our model is robust and really meaningful, we took a number of steps at test time $T_{\textrm{test}}=30$ greater than $T_{\textrm{train}}=15$. We see on Figure \ref{fig:poisson-generalization} that the performance are slightly better than for $T_{\textrm{test}}=15$, which shows that our model is robust to the number of iterations.

\subsection{Multivariate Distribution}
    This simulator is a combination of canonical distributions, aimed at showing that our architecture has enough capacity to learn an optimization procedure valid for parameters with very different impact on the samples. We also took a number of steps at test time $T_{\textrm{test}}=30$ greater than $T_{\textrm{train}}=15$. It can be observed from the left part of Figure \ref{fig:nmd} that the RMSE doesn't increase after the iteration 15 which shows that the learned procedure hasn't overfitted on the number of steps. This figure also shows that the procedure converges in both mean and variance.
    
    It shows that the architecture has enough capacity to learn an update procedure which eventually converges to a value close to the MLE, even in the case where the parameters have very different effects on the generated data.

\subsection{Linear Regression}
\begin{figure}[H]
    \centering
        \includegraphics[width=.5\textwidth]{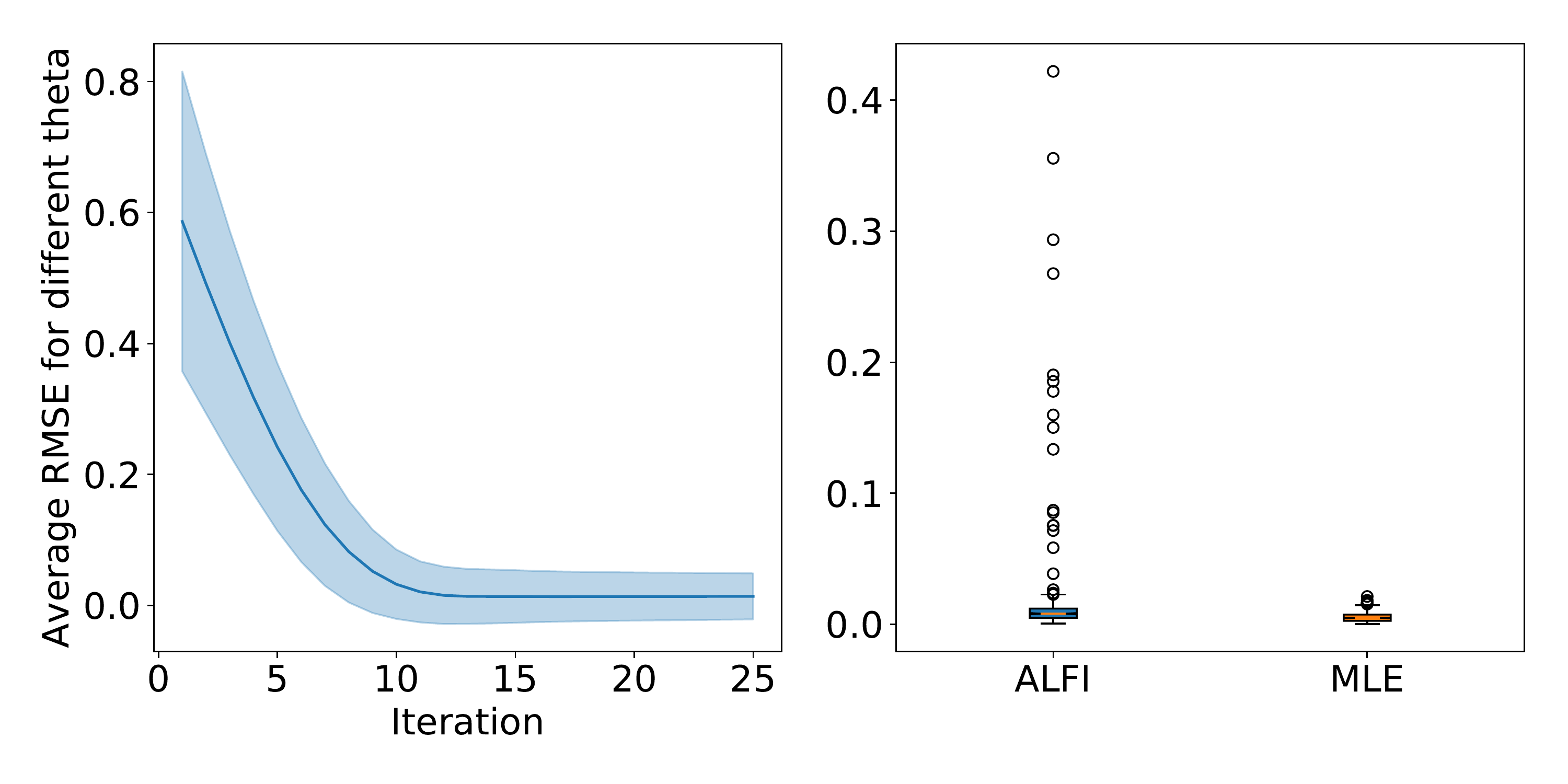}
        \caption{Results for the Linear Regression simulator. \textbf{(Left)} Fast convergence to a small average RMSE. \textbf{(Right)} ALFI gives numerous outliers but for the other points the performance of ALFI is comparable to the MLE. }
        \label{fig:lr}
\end{figure}

Figure \ref{fig:lr} presents the results for the linear regression simulator. We also took a number of steps at test time $T_{\textrm{test}}=30$ greater than $T_{\textrm{train}}=15$. It can be observed from the boxplot that, in average, ALFI has performance comparable with the MLE. However, we can notice the difficulty to estimate precisely the value of parameters for few cases. 

The left sub-figure shows that the RMSE quickly converges in 15 iterations and then is stable with a slight variance reduction during the 10 last iteration. 

\end{document}